\documentclass{article}

\usepackage{
    amsmath,  % For maths
    amssymb,  % For \mathbb
    array,  % For tables
    booktabs,  % For tables
    collas2025_conference,  % For CoLLAs formatting
    enumitem,  % For lists
    graphicx,  % For figures
    microtype,  % For better typesetting
    ragged2e,  % For tables
    times,  % For Times New Roman font
    xcolor,  % For colors
}
\usepackage[hyperfootnotes=false]{hyperref}
\usepackage[capitalize,nameinlink,noabbrev]{cleveref}
\usepackage[font=small,labelfont=bf,labelsep=quad,justification=justified,singlelinecheck=false]{caption}
\usepackage[font=small,labelfont=normalfont,justification=centering,singlelinecheck=false]{subcaption}

\collasfinalcopy

\hypersetup{
    colorlinks,
    linkcolor={blue!50!black},
    citecolor={blue!50!black},
    urlcolor={blue!50!black},
}

\creflabelformat{equation}{#2\textup{#1}#3}

\newcommand{\entropy}[1]{\mathrm{H} \! \left[ #1 \right]}
\newcommand{\expectation}[2]{\mathbb{E}_{#1} \! \left[ #2 \right]}
\newcommand{\kldivergence}[2]{\mathrm{KL} \! \left[ #1 \, \| \, #2 \right]}

\newcommand{\shorturl}[1]{\href{https://#1}{\nolinkurl{#1}}}

\title{Prediction-Oriented Subsampling from Data Streams}

\author{%
Benedetta L. Mussati\thanks{Equal contribution. Correspondence to \texttt{blmussati@robots.ox.ac.uk} and \texttt{freddie@robots.ox.ac.uk}.} \\
University of Oxford
\And
Freddie Bickford Smith$^*$ \\
University of Oxford
\And
Tom Rainforth \\
University of Oxford
\And
Stephen Roberts \\
University of Oxford \\
\& Mind Foundry
}

\begin{document}

\raggedbottom

\maketitle

\vspace{-10pt}

\begin{abstract}
    Data is often generated in streams, with new observations arriving over time.
A key challenge for learning models from data streams is capturing relevant information while keeping computational costs manageable.
We explore intelligent data subsampling for offline learning, and argue for an information-theoretic method centred on reducing uncertainty in downstream predictions of interest.
Empirically, we demonstrate that this prediction-oriented approach performs better than a previously proposed information-theoretic technique on two widely studied problems.
At the same time, we highlight that reliably achieving strong performance in practice requires careful model design.
\end{abstract}

\section{Introduction}\label{sec:introduction}

While data is often treated as a static resource, in many cases it is produced in an ongoing stream: fleets of self-driving cars continuously record the road \citep{sun2020scalability}; markets produce ever-changing economic signals \citep{einav2014economics}; weather sensors give regular readouts of multitudes of environmental variables \citep{lam2023learning}.
Learning from a data stream can sometimes be straightforward, such as when performing exact Bayesian inference with an appropriate model \citep{zellner1988optimal}.
Yet more generally it has been a challenging problem to make progress on \citep{prabhu2020gdumb}.

We suggest better methods for intelligent data subsampling could be crucial for more effectively working with data streams.
Online learning is conceptually appealing, but practical implementations typically fail to match the theoretical ideal that motivates them \citep{nguyen2018variational,pan2020continual,ritter2018online,rudner2022continual}.
Meanwhile an idea that has seen significant success in the continual learning literature is that of ``replay'' \citep{lin1992selfimproving} or ``rehearsal'' \citep{ratcliff1990connectionist,robins1995catastrophic} from a fixed-size data store constructed by subsampling from the data stream \citep{vandeven2024continual}.
Naive uniform-random subsampling \citep{chaudhry2019continual,rolnick2019experience,prabhu2020gdumb} can lead to significant information loss, since there is nothing to stop the data store from being filled with uninformative examples.
But we know that more sophisticated approaches can allow us to discard most of the examples in a dataset and still retain all or most of the relevant information \citep{angelova2004data,lewis1994sequential,schohn2000less}.

Inspired by recent advances in active learning \citep{bickfordsmith2023prediction,bickfordsmith2024making}, we argue that subsampling should focus on information gain in the downstream predictions that we want to make, explicitly accounting for the input distribution of interest.
We contrast this prediction-oriented approach with a notable existing method \citep{sun2022information}, which uses information-theoretic concepts but does not directly measure information gain and, insofar as it relates to information gain, it does so with respect to model parameters rather than the predictions we ultimately care about.

\looseness=-1
In empirical evaluations on Split MNIST and Split CIFAR-10 \citep{zenke2017continual}, two data streams popular in the continual learning literature \citep{vandeven2024continual}, we find improved performance relative to the aforementioned baseline method.
Importantly we also demonstrate the sensitivity of information-theoretic subsampling to how we construct our predictive models, clarifying the scope for new best practices in model design to help improve subsampling.
\section{Why learning from a data stream is difficult}\label{sec:data_problem}

We can think of a stream of labelled data as producing $n$ input-label pairs, $(x_{t,i},y_{t,i})_{i=1}^n$, at each time step, $t$:
\begin{align*}
    (x_{t,i},y_{t,i}) \sim p_t(x,y) = p_t(x|y)p_t(y) = p_t(y|x)p_t(x)
    .
\end{align*}
We begin by clarifying the technical challenge of learning from a data stream.
With reference to the continual learning literature, we suggest that the challenge is best thought of as a progression of three issues, which can be exacerbated by nonstationarity (time-dependence) in data generation, but remain even if the data-generating process is stationary.

\subsection{Nonstationarity is an important but unnecessary factor}

A nonstationary data stream is one for which the data-generating distribution changes with time:
\begin{align*}
    p_t(x,y) \neq p_{t'}(x,y) \mathrm{\ for\ } t \neq t'
    .
\end{align*}
Nonstationarity can come in different forms, with four special cases being
\begin{align*}
    p_t(x,y) &= p_0(y|x)p_t(x) \qquad \mathrm{(marginal\ input\ shift)}
    \\
    p_t(x,y) &= p_0(x|y)p_t(y) \qquad \mathrm{(marginal\ label\ shift)}
    \\
    p_t(x,y) &= p_t(x|y)p_0(y) \qquad \mathrm{(conditional\ input\ shift)}
    \\
    p_t(x,y) &= p_t(y|x)p_0(x) \qquad \mathrm{(conditional\ label\ shift)}
\end{align*}
where $p_0$ denotes a stationary distribution.
Pairs of these special cases can coincide.
Marginal input shift and conditional label shift are sometimes called covariate shift and concept drift respectively \citep{farquhar2022what}.

Many of the data streams used for evaluations in the continual learning literature are nonstationary: streams like Split MNIST and Split CIFAR-10~\citep{zenke2017continual} can be understood as implementing marginal label shift, while streams like Permuted MNIST \citep{goodfellow2015empirical} and CLEAR \citep{lin2021clear} implement conditional input shift.
This, combined with how work on continual learning has often been motivated in the literature \citep{aljundi2019gradient,lin2021clear,lopezpaz2017gradient,prabhu2024random,rolnick2019experience,vandeven2024continual,zenke2017continual}, can give the impression that nonstationarity is the defining reason why data streams pose a technical challenge.

Nonstationarity is an important and interesting factor in many practical problems, and we use nonstationary data streams in our empirical evaluations for consistency with previous work.
But we stress that the difficulty of working with data streams does not depend on nonstationarity: the issues we highlight here hold even for stationary data streams.

\subsection{Online learning is appealing in theory but limited in practice}\label{sec:online_learning}

Having discussed the nature of data streams, we now turn to the problem of learning from them.
In particular we consider learning a predictive model of the form $p_\phi(y|x)$ where $\phi$ is an index over the model class.
This model can contain some stochastic parameters, $\theta \sim p_\phi(\theta)$, defined such that $p_\phi(y_{1:k}|x_{1:k}) = \expectation{p_\phi(\theta)}{p_\phi(y_{1:k}|x_{1:k},\theta)}$ and $p_\phi(y_{1:k}|x_{1:k},\theta)=p_\phi(y_1|x_1,\theta)p_\phi(y_2|x_2,\theta)\ldots p_\phi(y_k|x_k,\theta)$.
We recover Bayesian models as a special case by letting $p_\phi(\theta)$ denote our beliefs over $\theta$ given observations (if any) and letting $p_\phi(y|x,\theta)$ be a fixed likelihood function.

Online learning is an elegant solution: simply update on new data as it arrives.
Information-theoretically optimal online learning losslessly compresses the data, such that we can discard the data after processing it.
Bayesian inference with a well-specified model matches this ideal: the posterior distribution is an optimal data representation \citep{zellner1988optimal}.
A corollary of this is the coherence property \citep{bissiri2016general} of Bayesian updating on new data, $d_{1:n}$:
\begin{align*}
    p_\phi(\theta|d_{1:n})
    =
    \frac{p_\phi(d_{1:n}|\theta)p_\phi(\theta)}{p_\phi(d_{1:n})}
    =
    \frac{p_\phi(d_b|\theta)p_\phi(\theta|d_a)}{p_\phi(d_b|d_a)}
    \quad \mathrm{for\ all\ } a \subset \{1, 2, \ldots, n\} \mathrm{\ and\ } b = \{1, 2, \ldots, n\} \setminus a
    .
\end{align*}
That is, we arrive at the same beliefs over the model parameters regardless of the order in which we receive the data.
This behaviour of Bayesian updating has inspired many approaches to online learning that take the form of repeated approximate inference, where an approximate posterior from one time step is used as the prior for the next time step.
Examples include online variants of Laplace approximation \citep{smola2003laplace} and variational inference \citep{sato2001online}.
The implicit hope is to avoid intractable inference \citep{murphy2022probabilistic} while maintaining the benefits of Bayesian updating.

We emphasise that online approximate inference does not behave the same as exact inference in the general case: using approximate posteriors in place of exact ones induces a loss of information \citep{turner2011two}.
This helps explain the suboptimal performance of recent implementations of this idea in the continual learning literature \citep{nguyen2018variational,pan2020continual,ritter2018online,rudner2022continual}.
Crucially it also justifies the widespread use of ``replay'' \citep{lin1992selfimproving} or ``rehearsal'' \citep{ratcliff1990connectionist,robins1995catastrophic} on examples subsampled from the data stream \citep{vandeven2024continual}.
Storing data is a straightforward way to retain information that would otherwise be lost in online learning.

\subsection{Storing all incoming data is not a general solution}\label{subsec:offlinelearning_alldata}

Given our conclusions from \Cref{sec:online_learning}, a natural idea is to simply store all the incoming data from a stream and use it for offline learning.
This solves the information-loss problem but introduces a new one: the cost of storing the data and training on it grows over time, which often untenable in practice.
Across the various alternatives to online learning that we consider, only one strategy does not grow in cost over time: offline learning on a fixed-size store of data subsampled from the stream (\Cref{tab:learning_strategies}).
This computationally efficient strategy stands out as a solution to explore.

\begin{table}[t]
    \centering
    \small
    \renewcommand{\arraystretch}{2}  % https://tex.stackexchange.com/a/394792
    \begin{tabular}{
        p{0.05cm}
        >{\raggedright\arraybackslash}p{2.1cm}
        >{\raggedright\arraybackslash}p{3.1cm}
        >{\raggedright\arraybackslash}p{3.2cm}
        >{\raggedright\arraybackslash}p{1.4cm}
        >{\raggedright\arraybackslash}p{2.1cm}
        >{\raggedright\arraybackslash}p{1.5cm}
    }
    \toprule
    & \textbf{Strategy} & \textbf{Online actions} & \textbf{Offline actions} & \textbf{Storage} & \textbf{Selection} & \textbf{Training}\\
    \midrule
    A & Online learning 
      & Train on $n$ examples 
      & None 
      & $0$
      & $0$
      & $C_\mathrm{train}(n)$
      \\
    \midrule
    B & Offline learning on all data 
      & Add $n$ examples to store 
      & Train on $nt$ examples 
      & $C_\mathrm{store}(nt)$ 
      & $0$
      & $C_\mathrm{train}(nt)/\tau$
      \\
    \midrule
    C & Offline learning on offline subset 
      & Add $n$ examples to store 
      & Select $m \le nt$ examples; \newline train on $m$ examples
      & $C_\mathrm{store}(nt)$ 
      & $C_\mathrm{select}(m,nt)/\tau$
      & $C_\mathrm{train}(m)/\tau$
      \\
    \midrule
    D & Offline learning on online subset (addition) 
      & Select $m \le n$ examples; \newline add $m$ examples to store 
      & Train on $mt$ examples 
      & $C_\mathrm{store}(mt)$ 
      & $C_\mathrm{select}(m,n)$ 
      & $C_\mathrm{train}(mt)/\tau$
      \\
    \midrule
    E & Offline learning on online subset (replacement) 
      & Select $m \le n$ examples; \newline replace $m$ examples in store
      & Train on $m$ examples
      & $C_\mathrm{store}(m)$ 
      & $C_\mathrm{select}(m,n)$ 
      & $C_\mathrm{train}(m)/\tau$
      \\
    \bottomrule
    \end{tabular}
    \caption{
        Five strategies for learning from a data stream pose different computational costs, with only A and E having costs that do not grow over time.
        We assume $n$ examples arrive at each time step, $t \in (1, 2, 3, \ldots)$.
        Online actions are performed at every step, while offline actions are performed every $\tau$ steps.
        Each strategy has three costs: $C_\mathrm{store}(N)$ is the cost of storing $N$ examples, $C_\mathrm{select}(M,N)$ is the cost of selecting a subset of $M \leq N$ examples, and $C_\mathrm{train}(N)$ is the cost of training on $N$ examples.
    }
    \label{tab:learning_strategies}
\end{table}

\subsection{Subsampling from a data stream needs careful consideration}

We have made a case that practical online learning can lead to high information loss and that offline learning on stored data can be a better idea, but keeping costs manageable requires only storing some of the data from a stream.
Now we turn to the question of how to subsample from a data stream and suggest why naive approaches can be problematic.

It is easy to conclude from past work on continual learning that subsampling from a data stream can be effective without much attention to how we do it.
For example, good results have been achieved from sampling uniformly at random from the data stream \citep{chaudhry2019continual,rolnick2019experience}, possibly balancing by class \citep{prabhu2020gdumb}.

We argue that these results are an artefact of the data streams popularly used in continual learning experiments, which are constructed from clean, curated datasets like MNIST \citep{lecun1998gradientbased} and CIFAR-10 \citep{krizhevsky2009learning}: the vast majority of the examples in these datasets are at least reasonably informative.
Work in the active-learning literature has shown that naive data acquisition performs significantly worse when aspects of messy real-world datasets are simulated \citep{bickfordsmith2023prediction,bickfordsmith2024making}.
This suggests that applying naive subsampling to real-world data streams could lead to substantial information loss, and that more careful approaches might be needed to mitigate this problem.
\section{Past efforts to use information theory fall short}\label{sec:mic_problem}

In \Cref{sec:data_problem} we established that the challenge of learning from a data stream can be thought of in terms of information loss.
Given this, a principled basis for reasoning about data subsampling is to explicitly measure the information that examples convey about a quantity of interest.
We highlight that the work by \cite{sun2022information} refers to rigorous information-theoretic principles that have seen success in other fields, but it ultimately settles on suboptimal heuristics.

Bayesian experimental design \citep{lindley1956measure,rainforth2024modern} serves as a rigorous framework for targeting useful data.
A core measure of data utility within this framework is information gain, which for a general quantity of interest, $\psi$, is given by the reduction in entropy \citep{shannon1948mathematical} in $\psi$ that results from observing new data, $(x,y)$:
\begin{align*}
    \mathrm{IG}_\psi(x,y)
    =
    \entropy{p_\phi(\psi)} - \entropy{p_\phi(\psi|x,y)}
\end{align*}
where $p_\phi(\psi)$ denotes prior beliefs over $\psi$ and $p_\phi(\psi|x,y) \propto p_\phi(y|x,\psi) p_\phi(\psi)$ denotes posterior beliefs given $(x,y)$.
Using this notion of data utility rigorously implements the idea that we want data to be informative with respect to $\psi$.

\citet{sun2022information} instead defined their ``memorable information criterion'' (MIC) based on intuitions of surprise and learnability, formalised as the difference of two negative log likelihoods weighted by a hyperparameter, $\eta$:
\begin{align*}
    \mathrm{MIC}(x,y,\eta)
    =
    \underbrace{-\log p_\phi(y'=y|x)}_\mathrm{surprise} + \eta \underbrace{\log p_\phi(y'=y|x,y)}_\mathrm{learnability}
    .
\end{align*}
For a high MIC value, we need a high negative log likelihood before observing $y$ (high surprise) and a low negative log likelihood after observing $y$ (high learnability).
\citet{sun2022information} were transparent in presenting MIC as only a heuristic inspired by information theory, and related it to a more well-established information-theoretic quantity, namely the Kullback-Leibler (KL) divergence between the parameter posterior, $p_\phi(\theta|x,y)$, and the parameter prior, $p_\phi(\theta)$:
\begin{align*}
    \kldivergence{p_\phi(\theta|x,y)}{p_\phi(\theta)}
    &=
    \expectation{p_\phi(\theta|x,y)}{\log \frac{p_\phi(\theta|x,y)}{p_\phi(\theta)}}
    \\
    &=
    \expectation{p_\phi(\theta|x,y)}{\log \frac{p_\phi(y'=y|x,\theta)}{p_\phi(y'=y|x)}} 
    \\
    &=
    -\log p_\phi(y'=y|x) + \expectation{p_\phi(\theta|x,y)}{\log p_\phi(y'=y|x,\theta)} 
    \\
    &\leq
    -\log p_\phi(y'=y|x) + \log \expectation{p_\phi(\theta|x,y)}{p_\phi(y'=y|x,\theta)} 
    \\
    &=
    \mathrm{MIC}(x,y,1) 
    .
\end{align*}
That is, MIC with $\eta=1$ is an upper bound on the KL divergence between $p_\phi(\theta|x,y)$ and $p_\phi(\theta)$.
Any looseness in this bound could lead MIC to differ from the KL divergence in the data it prioritises.
More importantly, we argue that this KL divergence is not even the most appropriate notion of data utility.
In machine learning we overwhelmingly care about predictions.
A principled information-theoretic approach should therefore target prediction-relevant information directly, and measuring data utility in terms of the model parameters is suboptimal \citep{bickfordsmith2023prediction,bickfordsmith2024making}.

In addition to this upper-bound perspective, we can interpret MIC with $\eta=1$ as the reduction in the model's predictive loss on $(x,y)$ that results from observing $(x,y)$.
The emphasis on prediction here arguably makes this interpretation a more compelling justification for using MIC than the parameter-oriented interpretation.
But maximising $\mathrm{MIC}(x,y,1)$ is problematic also from a predictive perspective.
First, it implicitly assumes that $x$ is the one and only input we want to make predictions on, whereas we generally predict on random samples from some task-dependent input distribution.
The particular $x$ on which we evaluate MIC does not even need to lie in the support of this distribution, so MIC can prioritise data with little relevance to the predictive task we care about.
Second, maximising $\mathrm{MIC}(x,y,1)$ implicitly assumes that we want to update our beliefs in order to assign more weight to the particular $y$ we have at hand. 
This can lead to worse overall predictive performance, for example if $y$ is drawn from an imperfect label source.
\section{Subsampling by maximising expected predictive information gain}\label{sec:method}

We now return to the idea of information gain and consider how it should be applied to the problem of data subsampling.
In particular we suggest it is important to focus on the downstream predictions that we want to make with our model, and we highlight conceptual challenges associated with subsampling data using knowledge of the labels.

\subsection{Focusing on information gain in downstream predictions}

Identifying the information-theoretically optimal measure of data utility involves returning to the idea of information gain (\Cref{sec:mic_problem}) with care to focus on the quantity we ultimately want to learn about.
In machine learning that quantity is typically the label, $y_*$, of a target input, $x_*$, leading us to predictive information gain:
\begin{align*}
    \mathrm{IG}_{y_*}(x,y,x_*) = \entropy{p_\phi(y_*|x_*)} - \entropy{p_\phi(y_*|x_*,x,y)}
    .
\end{align*}
Because in the general case we do not know which inputs we will want to predict on, $x_*$ can be thought of as a random variable distributed according to some target input distribution: $x_* \sim p_*(x_*)$.
In the context of label acquisition, $y$ is also unknown when making a data-acquisition decision so it too is treated as a random variable.
Taking an expectation over both $x_*$ and $y$ leads to the expected predictive information gain (EPIG; \citealp{bickfordsmith2023prediction}):
\begin{align*}
    \mathrm{EPIG}(x) = \expectation{p_*(x_*)p_\phi(y|x)}{\mathrm{IG}_{y_*}(x,y,x_*)}
    .
\end{align*}
In the context of subsampling a labelled dataset, each $x$ is associated with a known $y$, and we do not have to treat $y$ as a random variable.
This makes it possible to use a label-aware notion of expected predictive information gain (LA-EPIG):
\begin{align*}
    \mathrm{LA}\text{-}\mathrm{EPIG}(x,y) = \expectation{p_*(x_*)}{\mathrm{IG}_{y_*}(x,y,x_*)}
    .
\end{align*}
LA-EPIG is evaluated for the particular observed $(x,y)$ within the dataset from which we are subsampling, whereas EPIG averages over the possible $y$ for a given $x$, with our model, $p_\phi(y|x)$, determining the weight on each possible $y$.
LA-EPIG was suggested as a possible data-subsampling objective by \citet{kirsch2022unifying}, who referred to it as PIG.

\subsection{Estimating LA-EPIG}\label{sec:la_epig_estimation}

Here we consider how to efficiently estimate LA-EPIG.
For simplicity we focus on the case where computing the entropy of a given predictive distribution is straightforward, such as when using a standard classification model with a categorical predictive distribution.
In this case the key challenge is obtaining $p_\phi(y_*|x_*,x,y)$, the predictive distribution after updating on a new input-label pair, $(x,y)$.
Explicit updating on each candidate $(x,y)$ can in some cases be cheap, such as when using a conjugate Bayesian model, but is typically expensive \citep{bickfordsmith2024making,bickfordsmith2025rethinking}.

We highlight that we can replace explicit updating with implicit updating through likelihood-based reweighting, since
\begin{align*}
    p_\phi(y_*|x_*,x,y) = \frac{p_\phi(y,y_*|x,x_*)}{p_\phi(y|x)}
    .
\end{align*}
This can be used for any model that produces nontrivial joint predictive distributions \citep{osband2022neural,osband2022evaluating,osband2023epistemic,wang2021beyond,wen2022from}.
In the common case of a stochastic model (\Cref{sec:online_learning}) whose predictive distribution cannot be computed analytically, we can use a Monte Carlo estimator:
\begin{align*}
    p_\phi(y_*|x_*,x,y)
    \approx
    \frac{\sum_{j=1}^K p_\phi(y'=y|x,\theta_j)p_\phi(y_*|x_*,\theta_j)}{\sum_{k=1}^K p_\phi(y'=y|x,\theta_k)}
    \quad \mathrm{where\ } \theta_j,\theta_k \sim p_\phi(\theta)
    .
\end{align*}
We can see this as add-one-in importance sampling \citep{fong2021conformal}.
The resulting LA-EPIG estimator is
\begin{align*}
    \mathrm{LA}\text{-}\mathrm{EPIG}(x,y) \approx
    \frac{1}{M} \sum_{i=1}^M
    \entropy{\frac{1}{K} \sum_{j=1}^K p_\phi(y_*|x_*^{(i)},\theta_j)}
    -
    \entropy{\frac{\sum_{j=1}^K p_\phi(y'=y|x,\theta_j)p_\phi(y_*|x_*^{(i)},\theta_j)}{\sum_{k=1}^K p_\phi(y'=y|x,\theta_k)}}
\end{align*}
where here we are also using Monte Carlo to approximate the expectation over $x_*$ using sampled $x_*^{(i)} \sim p_*(x_*)$.
There is a large degree of flexibility in how this target-input sampling can be implemented \citep{bickfordsmith2023prediction}.

\subsection{Using knowledge of labels in subsampling}\label{sec:la_epig_issues}

Given both EPIG and LA-EPIG can be used for subsampling from a labelled dataset, we now consider which should be preferred.
LA-EPIG initially appeals because it measures the utility of the labels we actually have at hand rather than measuring expected utility over possible labels and, as discussed in \Cref{sec:la_epig_estimation}, it can be computationally cheaper to estimate.
However, as we will explain, it actually suffers from significant theoretical and practical issues.

We first consider the problem with LA-EPIG from a frequentist perspective.
If $\ell$ is a loss function, $p_{\mathrm{true}}(x,y)$ is a true data-generating distribution and $f$ is a predictive model, then risk minimisation identifies the optimal predictive model as $f^* = \arg\min_{f} \expectation{p_{\mathrm{true}}(x,y)}{\ell(f(x),y)}$.
Now if we assume a fully unrestricted form on $f$, then by standard results \citep{hastie2009elements} the optimal predictor is given by piecewise solving the optimisation for each possible $x$, namely $f^*(x) = \arg\min_{f(x)} \expectation{p_{\mathrm{true}}(y|x)}{\ell(f(x),y)}$, and $f^*$ has no dependency on $p_{\mathrm{true}}(x)$.
This is a key element in the justification for any kind of active learning that makes targeted choices for the $x$ used for model training \citep{farquhar2021statistical}.
(Notably we often use models that are heavily overparameterised with respect to the dataset size, such that the assumption that $f$ is unrestricted is reasonable.)
However, these justifying arguments for active learning are violated once we take any kind of active role in choosing input-label pairs rather than inputs alone: then we are no longer sampling labels according to $p_{\mathrm{true}}(y|x)$, and the optimal predictive model for data drawn in this way will be inherently different to that which optimises the true risk, $\expectation{p_{\mathrm{true}}(x,y)}{\ell(f(x),y)}$, regardless of model capacity.
Thus LA-EPIG will lead to collecting datasets that no longer properly represent the predictive problem we wish to solve.

We can also criticise LA-EPIG from a Bayesian perspective.
The likelihood function, $p_{\phi}(y|x,\theta)$, in a Bayesian model is usually constructed to model some true data-generating process under a given hypothesis, $\theta$.
If we directly choose training examples, $(x,y)$, by maximising LA-EPIG then a likelihood function constructed in this way is misspecified \citep{kleijn2006misspecification}: it does not incorporate the data-selection procedure and so does not reflect the true data-generating process.
In principle this misspecification is something that can be addressed.
For example, we could implicitly construct a valid likelihood function to update beliefs in this setting by simulating the whole source dataset, taking $(x,y)$ as the pair in the simulated source dataset with the highest LA-EPIG score to implicitly define $p_{\phi}(x,y|\theta)$, then disintegrating this joint at our given $x$ to define $p_{\phi}(y|x,\theta)$.
This is difficult to implement in practice, so in reality we expect to update our beliefs using a misspecified model, which can cause issues \citep{watson2016approximate}.

The problem with LA-EPIG can additionally be understood in terms of the bias in the data it prioritises: as we will discuss in \Cref{sec:visual_demo}, it can favour data that reinforces the model's existing predictive beliefs.
This is because we intuitively want our uncertainty to decrease less if we make new observations that are at odds with our previous beliefs than if they agree with our prior beliefs.  
For example, if we have a datapoint whose label is distinct to those of nearby points, our model must either allow for output noise or high frequency fluctuations to accommodate it, such that incorporating it will tend to raise its predictive uncertainty.
Thus, if our model is appropriately set up, it will generally prefer the datapoint whose label agrees with the current beliefs over the one that disagrees.

Overall we believe these issues justify EPIG as a preferable default over LA-EPIG.
Our empirical investigation in \Cref{sec:experiments} also support this recommendation: EPIG consistently produces stronger performance than LA-EPIG.
\section{Visual demonstration}\label{sec:visual_demo}

\begin{figure*}[t]
    \vspace{-10pt}
    \centering
    \includegraphics[width=\linewidth]{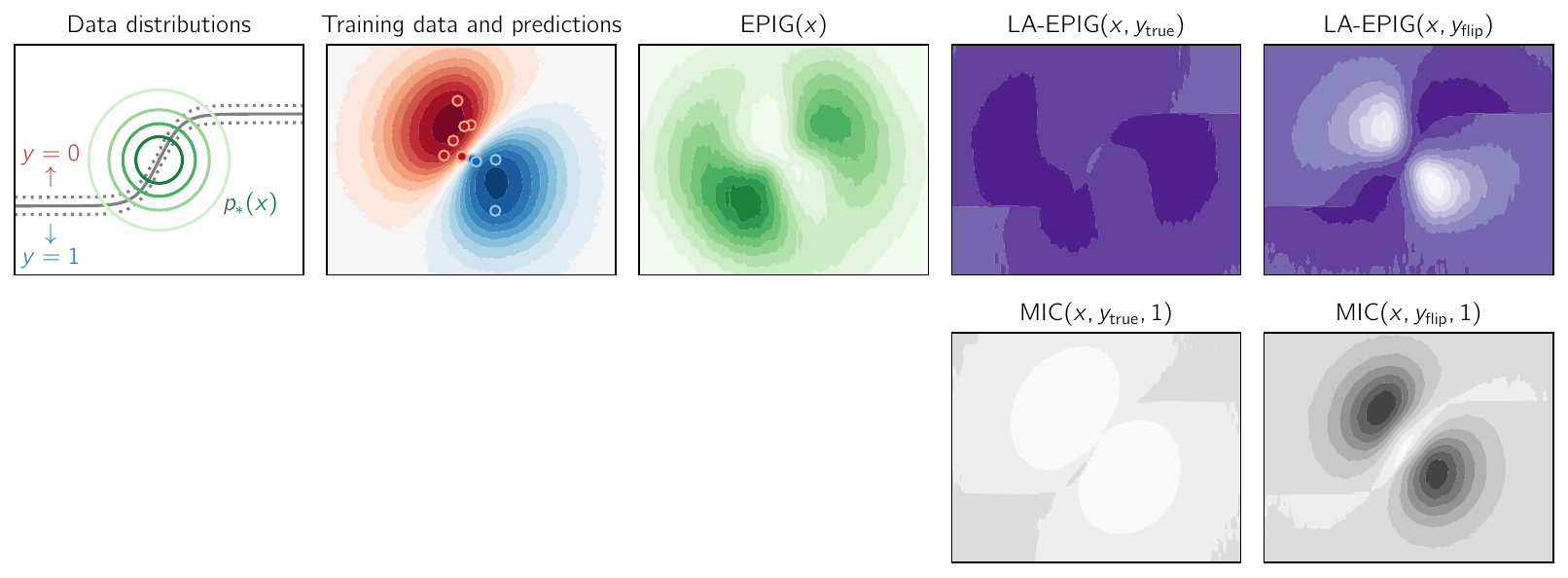}
    \caption{
        Data prioritisation (darker shades indicate higher values) varies substantially between the expected predictive information gain (EPIG; \citealp{bickfordsmith2023prediction}), a label-aware variant (LA-EPIG) and the ``memorable information criterion'' (MIC;  \citealp{sun2022information}).
        LA-EPIG and MIC are evaluated on ``true'' labels, $y_\mathrm{true} = \arg\max_{y} p_\mathrm{true}(y|x)$, as well as ``flipped'' labels, $y_\mathrm{flip} = 1 - y_\mathrm{true}$.
    }
    \label{fig:two_bells_heatmaps}
\end{figure*}

To provide some intuition about how different methods prioritise data, we visualise MIC (with $\eta=1$), EPIG and LA-EPIG on a simple binary-classification problem introduced by \citet{bickfordsmith2023prediction}, inheriting their implementation details (\Cref{fig:two_bells_heatmaps}).
The task of interest is to predict $y_*|x_*$ for $x_* \sim p_*(x_*)$.
We use a Gaussian-process classifier and a variational approximation to the Bayesian posterior over latent-function values given training data.

Our plots help explain how MIC can be suboptimal.
MIC is large for an input-label pair, $(x,y)$, if the model currently places low predictive weight on $y$ and places high weight on $y$ after updating on $(x,y)$.
In our plots we see that this translates to MIC favouring local changes in the model's predictions without regard for the inputs we are targeting, $x_* \sim p_*(x_*)$.
When evaluated on true labels, MIC prioritises data that is simply far away from the training data, which introduces a risk of gathering data of limited relevance to the task of interest, much like the pathology of BALD \citep{houlsby2011bayesian} demonstrated by \citet{bickfordsmith2023prediction}.
For flipped labels, MIC prioritises data that contradicts the model's predictions as much as possible, undermining the training data we already have.

Comparing LA-EPIG against MIC reveals a notable contrast in behaviour.
LA-EPIG is high for a given input-label pair, $(x,y)$, if updating the model on $(x,y)$ leads to a large expected reduction in predictive uncertainty on the target input, $x_*$, where the expectation is with respect to the target input distribution, $p_*(x_*)$.
Explicitly accounting for $p_*(x_*)$ allows LA-EPIG to avoid MIC's pitfall of prioritising data that has limited relevance to the task of interest.
The $(x,y)$ pairs it favours are separate from the training data but still have relatively high density under the target input distribution.

At the same time, \Cref{fig:two_bells_heatmaps} shows how LA-EPIG can cause issues.
Maximising LA-EPIG leads us to select whichever input-label pair, $(x,y)$, produces the lowest uncertainty in $p_\phi(y_*|x_*,x,y)$ on average over different $x_*$.
This biases us away from data that disagrees with the model's existing beliefs.
While we might think of objectives like LA-EPIG as necessarily promoting changes in the model's beliefs, the objectives are always computed using the beliefs themselves, which can make truth-seeking difficult.
This failure mode is not unique to LA-EPIG, but LA-EPIG's use of specific labels, rather than probabilistic predictions, increases the chance that it will occur.

\section{Performance evaluation}\label{sec:experiments}

\begin{figure*}[t]
    \centering
    \begin{subfigure}[b]{\textwidth}
        \centering
        \includegraphics[width=\linewidth]{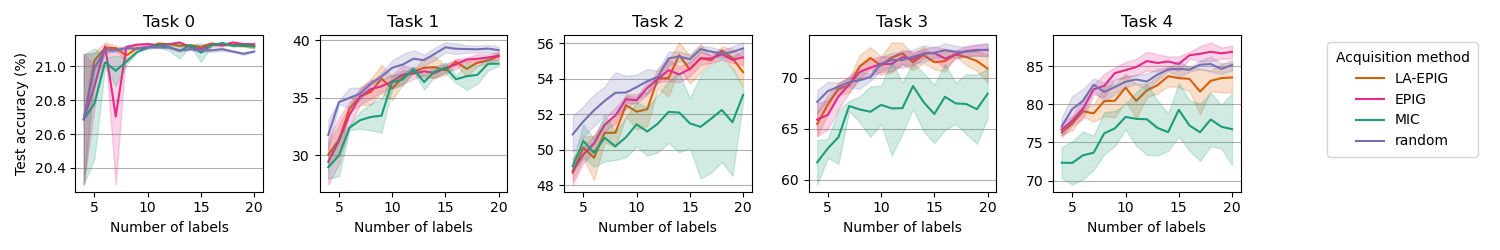}
        \caption{Dropout CNN on Split MNIST}
        \label{fig:mnist_cnn_m100}
    \end{subfigure}
    \\
    \vspace{10pt}
    \begin{subfigure}{\textwidth}
        \centering
        \includegraphics[width=\linewidth]{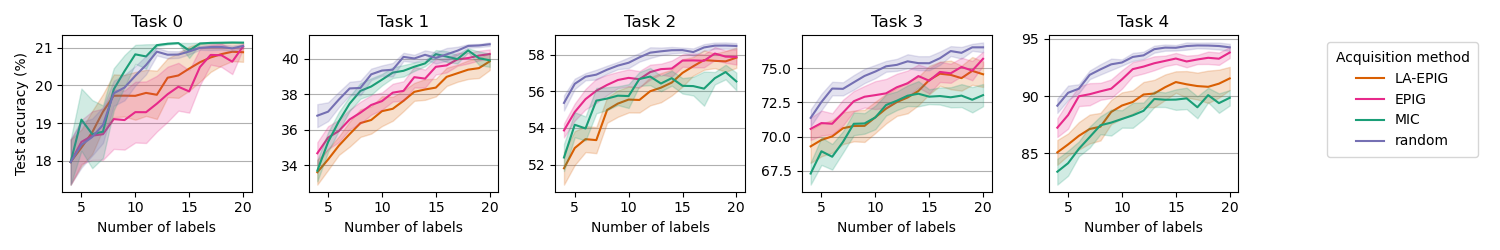}
        \caption{SimCLR encoder + dropout MLP on Split MNIST}
        \label{fig:mnist_simclr_mlp_m100}
    \end{subfigure}
    \caption{
        Capitalising on unlabelled data leads to stronger baseline predictive performance, but that does not necessarily translate to better data subsampling.
        Here a semi-supervised model (comprising an unsupervised encoder and a supervised prediction head) outperforms a fully supervised model when both models are trained on randomly subsampled data.
        At the same time, the benefit of intelligent subsampling (with EPIG) that we see for the fully supervised model does not hold for the semi-supervised model.
    }
    \label{fig:supervised_vs_semisupervised}
\end{figure*}

Now we investigate the practical use of subsampling methods to support learning from data streams.
In particular, we consider using a given data stream as a source of examples from which we can construct a dataset for offline learning (\Cref{sec:data_problem}), where the subsampling method determines which examples from the stream are stored.
Code for reproducing our results is available at \shorturl{github.com/blmussati/subsampling-data-streams}.

\subsection{Setup}

\paragraph{Data streams}

We focus on two streams: Split MNIST and Split CIFAR-10 \citep{zenke2017continual}.
Each of these is based on a ten-way classification problem, with performance measured by accuracy (based on a zero-one loss) on the standard MNIST \citep{lecun1998gradientbased} and CIFAR-10 \citep{krizhevsky2009learning} test sets.
The training data arrives in the form of a stream over five time steps (or ``tasks'').
At each step we observe $n$ examples from two classes not observed so far.

\paragraph{Subsampling}

We assume a data store of size $m=100$ and populate the store by adding $m/5$ subsampled examples at each of the five time steps (more sophisticated approaches are possible but we prioritise experimental simplicity).
For subsampling we consider an iterative approach comparable to standard pool-based active learning \citep{lewis1994sequential}: we build up our training set one example at a time, with subsampling decisions interleaved with model updating.
Each subsampling decision is an index in $(1,2,\ldots,n)$ for inclusion in the $m/5$ examples for the given time step, and is produced either by maximising a subsampling objective (MIC, LA-EPIG or EPIG) or by uniform-random sampling.

\paragraph{Models}

Inspired by recent work highlighting the importance of model design in information-theoretic data acquisition \citep{bickfordsmith2024making}, this is a key variable we explore in our experiments.
We consider four models.
One is fully supervised: the two-convolutional-layer model from \citet{kirsch2019batchbald} with their dropout rate of 0.5.
The remaining three models are semi-supervised, each with a fixed, unsupervised encoder that incorporates information from unlabelled data and a trainable, supervised prediction head that incorporates information from labelled data: (1) a SimCLR \citep{chen2020simple} encoder with a multilayer-perceptron (MLP) prediction head; (2) a DINOv2 \citep{oquab2024dinov2} encoder with a random-forest prediction head; and (3) a DINOv2 encoder with an MLP prediction head.
The SimCLR encoder is trained on unlabelled images from the domain of interest (ie, MNIST images for Split MNIST), while the DINOv2 encoder is trained on unlabelled images from a separate domain. 
The MLP prediction head has three hidden layers with 128 units, ReLU nonlinearities, dropout at a rate of 0.1, and dense connections between all layers.

\paragraph{Neural-network training}

We use a training objective that combines the negative log likelihood (NLL) of the model parameters on the training data with a penalty on the Euclidean norm of the parameter vector.
We run up to 100,000 steps of gradient descent on the training objective with a learning rate of 0.01, recording the NLL on a small validation set after each gradient step, then restoring the parameter configuration corresponding to the lowest validation NLL.

\begin{figure*}[t]
    \centering
    \begin{subfigure}{\textwidth}
        \centering
        \includegraphics[width=\linewidth]{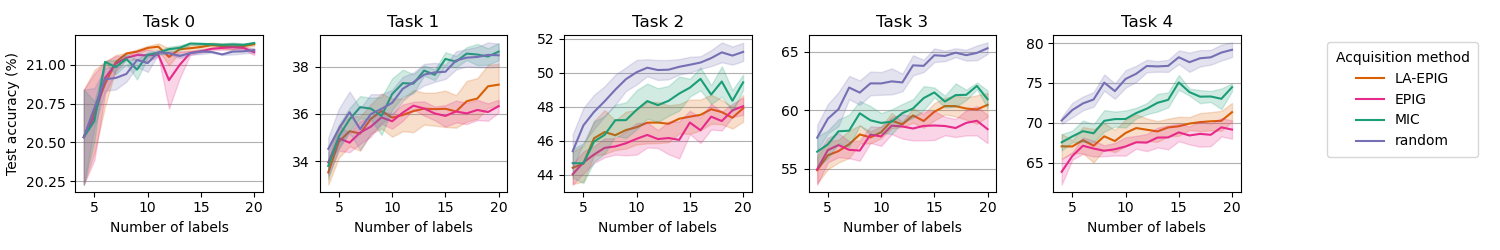}
        \caption{DINOv2 encoder + dropout MLP on Split MNIST}
        \label{fig:mnist_dino_mlp_m100}
    \end{subfigure}
    \\
    \vspace{10pt}
    \begin{subfigure}{\textwidth}
        \centering
        \includegraphics[width=\linewidth]{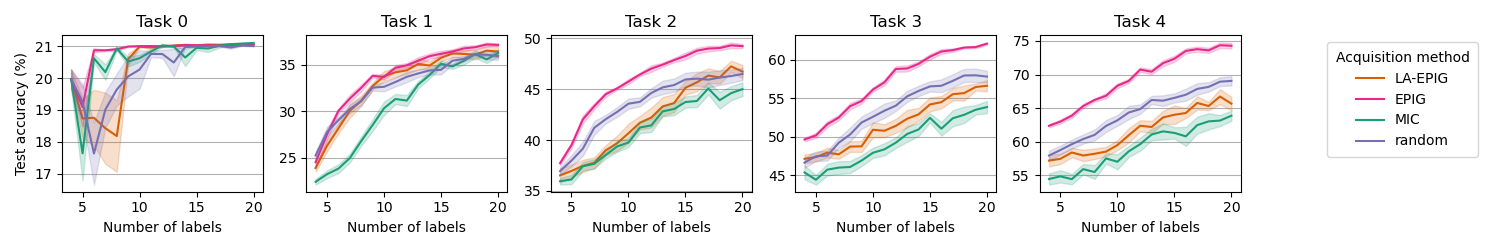}
        \caption{DINOv2 encoder + random forest on Split MNIST}
        \label{fig:mnist_dino_rf_m100}
    \end{subfigure}
    \caption{
        The benefit of intelligent subsampling for training a semi-supervised model depends strongly on the construction of the model.
        Here intelligent subsampling has an adverse effect on predictive performance for a model that uses a dropout MLP for its prediction head, but EPIG-based subsampling has a positive effect for a model that uses a random forest in place of the dropout MLP.
    }
    \label{fig:rf_vs_mlp_mnist}
\end{figure*}

\begin{figure*}[t]
    \centering
    \begin{subfigure}{\textwidth}
        \centering
        \includegraphics[width=\linewidth]{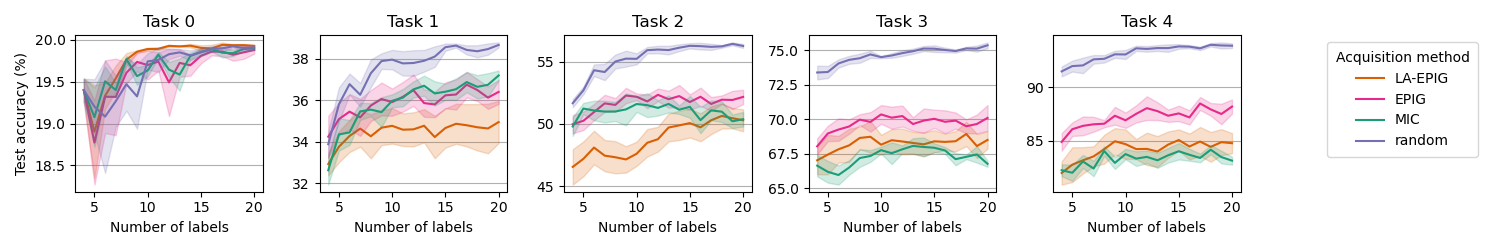}
        \caption{DINOv2 encoder + dropout MLP on Split CIFAR-10}
    \end{subfigure}
    \\
    \vspace{10pt}
    \begin{subfigure}{\textwidth}
        \centering
        \includegraphics[width=\linewidth]{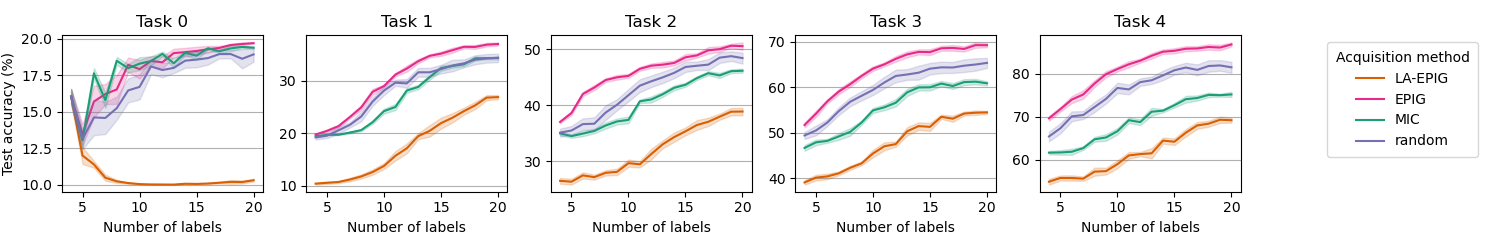}
        \caption{DINOv2 encoder + random forest on Split CIFAR-10}
    \end{subfigure}
    \caption{
        Aligning with results on Split MNIST (\Cref{fig:rf_vs_mlp_mnist}), the performance of intelligent subsampling on Split CIFAR-10 shows a strong dependence on model construction.
        Here EPIG-based subsampling is beneficial for one model but not the other.
    }
    \label{fig:rf_vs_mlp_cifar10}
\end{figure*}

\subsection{Findings}

First, focusing on Split MNIST, we explore the effect of switching from a fully supervised model to a semi-supervised model that incorporates information from unlabelled data (\Cref{fig:supervised_vs_semisupervised}).
We find that the semi-supervised setup gives better baseline test accuracy, but intelligent subsampling has a negative effect in the semi-supervised model we consider.

To better understand this finding, we revisit the same problem with two different semi-supervised models (\Cref{fig:rf_vs_mlp_mnist}).
Switching from an encoder that has been trained in the specific domain of interest (\Cref{fig:mnist_simclr_mlp_m100}) to an encoder trained in a separate domain (\Cref{fig:mnist_dino_mlp_m100}) leads to a fall in baseline test accuracy and a similar failure of intelligent subsampling to produce any benefit over random subsampling.
Notably, though, we find that using an alternative prediction head, specifically a random forest (\Cref{fig:mnist_dino_rf_m100}), allows for a substantial benefit from intelligent subsampling using EPIG.

Translating this pair of semi-supervised models to Split CIFAR-10 (\Cref{fig:rf_vs_mlp_cifar10}), we find a strikingly similar qualitative result.
EPIG-based subsampling produces a clear performance boost over random subsampling when using a random forest as the prediction head, but it has a negative effect when using a dropout MLP as the prediction head.

Overall our results show that intelligent subsampling using EPIG can be useful but only with an appropriately constructed model.
While aiming to reduce uncertainty in downstream predictions (ie, maximising EPIG) is a principled approach, successfully doing so in practice requires care.
Forecasting changes in uncertainty critically relies on more than a model's predictive performance in terms of classification accuracy (eg, based on a zero-one loss): it requires appropriate predictive correlations \citep{osband2022neural,osband2022evaluating,wang2021beyond}.
This requirement is often not met by Bayesian deep learning \citep{bickfordsmith2024making}, helping to explain the failures we see with dropout MLPs in this work.
\section{Related work}

Data subsampling---also referred to as coreset selection, data curation and data filtering, among other names---has been shown to allow a considerable reduction in dataset size while retaining most or all of the relevant information \citep{angelova2004data,lewis1994sequential,schohn2000less}.
A recently proposed subsampling objective with a close connection to MIC (\Cref{sec:mic_problem}) is the reducible holdout loss (RHO-LOSS; \citealp{mindermann2022prioritized}):
\begin{align*}
    \mathrm{RHO}\text{-}\mathrm{LOSS}(x,y) = -\log p_\phi(y'=y|x) + \log p_\lambda(y'=y|x)
\end{align*}
where $p_\lambda$ denotes an auxiliary model trained on a holdout dataset, $(x_{1:k}^\mathrm{hold},y_{1:k}^\mathrm{hold})$, with minimal loss on this holdout dataset being the motivating goal.
We can view MIC with $\eta=1$ as an instance of RHO-LOSS where the holdout dataset is defined to simply comprise the input-label pair being considered for subsampling and the auxiliary model is produced by a Bayesian update of the main model: this gives $p_\lambda(y'=y|x) = p_\phi(y'=y|x,y)$.
RHO-LOSS also has a connection with EPIG in its focus on downstream predictions, but the two methods differ in what they target (loss vs uncertainty) and their corresponding requirement for labels (EPIG only requires samples of unlabelled inputs).

Subsampling has seen successful use in continual learning as part of ``replay'' \citep{lin1992selfimproving} and ``rehearsal'' \citep{ratcliff1990connectionist,robins1995catastrophic} techniques, which involve storing data from a data stream and using it for model training \citep{vandeven2024continual}.
Aside from uniform-random subsampling \citep{chaudhry2019continual,rolnick2019experience,prabhu2020gdumb} and information-theoretic methods, a notable group of methods are those that use gradients of a loss function with respect to model parameters.
This group includes gradient-based sample selection (GSS; \citealp{aljundi2019gradient}) and online coreset selection (OCS; \citealp{yoon2022online}), which both use notions of gradient diversity to inform subsampling.
\section{Conclusion}

We have argued that prediction-oriented, information-theoretic data subsampling could be a valuable tool for learning from data streams.
In light of practically limited techniques for online learning, subsampling provides an effective way to mitigate information loss over time, and information theory helps identify a rigorous subsampling objective for the context of machine learning, where our ultimate aim is typically to make downstream predictions with our model.

Our empirical results make clear that a critical challenge in implementing this approach in practice is the need for models that support effective uncertainty estimation.
Combining fixed, deterministic, unsupervised encoders with trainable, stochastic, supervised prediction heads can work well \citep{bickfordsmith2024making}, but we still lack best practices for constructing effective neural-network prediction heads.
Future work should aim to establish best practices, building on promising methodological developments proposed in recent years, for example by \citet{osband2023fine,osband2023epistemic}.

Alongside efforts to improve models, work could be done to implement information-theoretic subsampling in a computationally efficient way.
The algorithmic approach used in our experiments is suboptimal: sequentially constructing a subsample requires many repetitions of model training, and in most cases it is unrealistic to assume we know the number of time steps we want to subsample across.
Promising areas for methods development include enabling batch subsampling and using notions of information loss to decide when to replace examples currently within a data store.
\section*{Acknowledgements}

Benedetta L. Mussati and Freddie Bickford Smith are supported by the UK EPSRC Centre for Doctoral Training in Autonomous Intelligent Machines and Systems (grants EP/S024050/1 and EP/L015897/1).
Benedetta L. Mussati is also supported by Mind Foundry Ltd.
Tom Rainforth is supported by the UK EPSRC grant EP/Y037200/1.

\clearpage

\bibliography{references.bib}

@article{oquab2024dinov2,
  author = {Oquab and Darcet and Moutakanni and Vo and Szafraniec and Khalidov and Fernandez and Haziza and Massa and {El-Nouby} and Assran and Ballas and Galuba and Howes and Huang and Li and Misra and Rabbat and Sharma and Synnaeve and Xu and Jegou and Mairal and Labatut and Joulin and Bojanowski},
  year         = {2024},
  title        = {{DINOv2}: learning robust visual features without supervision},
  journal      = {Transactions on Machine Learning Research},
}

@article{watson2016approximate,
    author  = {Watson and Holmes},
    year    = {2016},
    title   = {Approximate models and robust decisions},
    journal = {Statistical Science},
}

@article{yoon2022online,
    author  = {Yoon and Madaan and Yang and Hwang},
    year    = {2022},
    title   = {Online coreset selection for rehearsal-based continual learning},
    journal = {International Conference on Learning Representations},
}

@article{turner2011two,
    author  = {Turner and Sahani},
    year    = {2011},
    title   = {Two problems with variational expectation maximisation for time series models},
    journal = {Bayesian Time Series Models},
}

@article{smola2003laplace,
    author  = {Smola and Vishwanathan and Eskin},
    year    = {2003},
    title   = {{Laplace} propagation},
    journal = {Conference on Neural Information Processing Systems},
}

@article{sato2001online,
    author  = {Sato},
    year    = {2001},
    title   = {Online model selection based on the variational {Bayes}},
    journal = {Neural Computation},
}

@article{zellner1988optimal,
    author  = {Zellner},
    year    = {1988},
    title   = {Optimal information processing and {Bayes's} theorem},
    journal = {The American Statistician},
}

@article{goodfellow2015empirical,
    author  = {Goodfellow and Mirza and Xiao and Courville and Bengio},
    year    = {2015},
    title   = {An empirical investigation of catastrophic forgetting in gradient-based neural networks},
    journal = {arXiv},
}

@article{prabhu2024random,
    author  = {Prabhu and Sinha and Kumaraguru and Torr and Sener and Dokania},
    year    = {2024},
    title   = {Random representations outperform online continually learned representations},
    journal = {Conference on Neural Information Processing Systems},
}

@article{lin2021clear,
    author  = {Lin and Shi and Pathak and Ramanan},
    year    = {2021},
    title   = {The {CLEAR} benchmark: continual learning on real-world imagery},
    journal = {Conference on Neural Information Processing Systems},
}

@article{lopezpaz2017gradient,
    author  = {Lopez-Paz and Ranzato},
    year    = {2017},
    title   = {Gradient episodic memory for continual learning},
    journal = {Conference on Neural Information Processing Systems},
}

@article{aljundi2019gradient,
    author  = {Aljundi and Lin and Goujaud and Bengio},
    year    = {2019},
    title   = {Gradient based sample selection for online continual learning},
    journal = {Conference on Neural Information Processing Systems},
}

@article{bissiri2016general,
    author  = {Bissiri and Holmes and Walker},
    year    = {2016},
    title   = {A general framework for updating belief distributions},
    journal = {Journal of the Royal Statistical Society: Series B (Statistical Methodology)},
}

@article{vandeven2024continual,
    author  = {{van de Ven} and Soures and Kudithipudi},
    year    = {2024},
    title   = {Continual learning and catastrophic forgetting},
    journal = {arXiv},
}

@article{lin1992selfimproving,
    author  = {Lin},
    year    = {1992},
    title   = {Self-improving reactive agents based on reinforcement learning, planning and teaching},
    journal = {Machine Learning},
}

@article{robins1995catastrophic,
    author  = {Robins},
    year    = {1995},
    title   = {Catastrophic forgetting, rehearsal and pseudorehearsal},
    journal = {Connection Science},
}

@article{ratcliff1990connectionist,
    author  = {Ratcliff},
    year    = {1990},
    title   = {Connectionist models of recognition memory: constraints imposed by learning and forgetting functions},
    journal = {Psychological Review},
}

@article{zenke2017continual,
    author  = {Zenke and Poole and Ganguli},
    year    = {2017},
    title   = {Continual learning through synaptic intelligence},
    journal = {International Conference on Machine Learning},
}

@article{chaudhry2019continual,
    author  = {Chaudhry and Rohrbach and Elhoseiny and Ajanthan and Dokania and Torr and Ranzato},
    year    = {2019},
    title   = {Continual learning with tiny episodic memories},
    journal = {International Conference on Machine Learning},
}

@article{rolnick2019experience,
    author  = {Rolnick and Ahuja and Schwarz and Lillicrap and Wayne},
    year    = {2019},
    title   = {Experience replay for continual learning},
    journal = {Conference on Neural Information Processing Systems},
}

@article{rudner2022continual,
    author  = {Rudner and {Bickford Smith} and Feng and Teh and Gal},
    year    = {2022},
    title   = {Continual learning via sequential function-space variational inference},
    journal = {International Conference on Machine Learning},
}

@article{ritter2018online,
    author  = {Ritter and Botev and Barber},
    year    = {2018},
    title   = {Online structured {Laplace} approximations for overcoming catastrophic forgetting},
    journal = {Conference on Neural Information Processing Systems},
}

@article{pan2020continual,
    author  = {Pan and Swaroop and Immer and Eschenhagen and Turner and Khan},
    year    = {2020},
    title   = {Continual deep learning by functional regularisation of memorable past},
    journal = {Conference on Neural Information Processing Systems},
}

@article{nguyen2018variational,
    author  = {Nguyen and Li and Bui and Turner},
    year    = {2018},
    title   = {Variational continual learning},
    journal = {International Conference on Learning Representations},
}

@article{prabhu2020gdumb,
    author  = {Prabhu and Torr and Dokania},
    year    = {2020},
    title   = {{GDumb}: a simple approach that questions our progress in continual learning},
    journal = {European Conference on Computer Vision},
}

@article{mindermann2022prioritized,
    author  = {Mindermann and Brauner and Razzak and Sharma and Kirsch and Xu and H{\"o}ltgen and Gomez and Morisot and Farquhar and Gal},
    year    = {2022},
    title   = {Prioritized training on points that are learnable, worth learning, and not yet learnt},
    journal = {International Conference on Machine Learning},
}

@article{sun2022information,
    author  = {Sun and Calandriello and Hu and Li and Titsias},
    year    = {2022},
    title   = {Information-theoretic online memory selection for continual learning},
    journal = {International Conference on Learning Representations},
}

@article{sun2020scalability,
    author = {Sun and Kretzschmar and Dotiwalla and Chouard and Patnaik and Tsui and Guo and Zhou and Chai and Caine and Vasudevan and Han and Ngiam and Zhao and Timofeev and Ettinger and Krivokon and Gao and Joshi and Zhao and Cheng and Zhang and Shlens and Chen and Anguelov},
    year = {2020},
    title = {Scalability in perception for autonomous driving: {Waymo} open dataset}, 
    journal = {Conference on Computer Vision and Pattern Recognition},
}

@article{lam2023learning,
    author = {Lam and Sanchez-Gonzalez and Willson and Wirnsberger and Fortunato and Alet and Ravuri and Ewalds and Eaton-Rosen and Hu and Merose and Hoyer and Holland and Vinyals and Stott and Pritzel and Mohamed and Battaglia},
    year = {2023},
    title = {Learning skillful medium-range global weather forecasting},
    journal = {Science},
}

@article{einav2014economics,
    author = {Einav and Levin},
    year = {2014},
    title = {Economics in the age of big data},
    journal = {Science},
}

@book{murphy2022probabilistic,
    author    = {Murphy},
    year      = {2022},
    title     = {Probabilistic Machine Learning: An Introduction},
    publisher = {MIT Press},
}

@article{bickfordsmith2023prediction,
    author  = {{Bickford Smith} and Kirsch and Farquhar and Gal and Foster and Rainforth},
    year    = {2023},
    title   = {Prediction-oriented {Bayesian} active learning},
    journal = {International Conference on Artificial Intelligence and Statistics},
}

@article{bickfordsmith2024making,
    author  = {{Bickford Smith} and Foster and Rainforth},
    year    = {2024},
    title   = {Making better use of unlabelled data in {Bayesian} active learning},
    journal = {International Conference on Artificial Intelligence and Statistics},
}

@article{bickfordsmith2025rethinking,
    author = {{Bickford Smith} and Kossen and Trollope and {van der Wilk} and Foster and Rainforth},
    year = {2025},
    title = {Rethinking aleatoric and epistemic uncertainty},
    journal = {International Conference on Machine Learning},
}

@article{farquhar2022what,
    author  = {Farquhar and Gal},
    year    = {2022},
    title   = {What ``out-of-distribution'' is and is not},
    journal = {Workshop on ``ML Safety'', Conference on Neural Information Processing Systems},
}

@article{chen2020simple,
    author  = {Chen and Kornblith and Norouzi and Hinton},
    year    = {2020},
    title   = {A simple framework for contrastive learning of visual representations},
    journal = {International Conference on Machine Learning},
}

@article{schohn2000less,
    author  = {Schohn and Cohn},
    year    = {2000},
    title   = {Less is more: active learning with support vector machines},
    journal = {International Conference on Machine Learning},
}

@article{farquhar2021statistical,
    author  = {Farquhar and Gal and Rainforth},
    year    = {2021},
    title   = {On statistical bias in active learning: how and when to fix it},
    journal = {International Conference on Learning Representations},
}

@mastersthesis{angelova2004data,
    author = {Angelova},
    year   = {2004},
    title  = {Data pruning},
    school = {California Institute of Technology},
}

@article{fong2021conformal,
    author  = {Fong and Holmes},
    year    = {2021},
    title   = {Conformal {Bayesian} computation},
    journal = {Conference on Neural Information Processing Systems},
}

@article{kleijn2006misspecification,
    author  = {Kleijn and {van der Vaart}},
    year    = {2006},
    title   = {Misspecification in infinite-dimensional {Bayesian} statistics},
    journal = {Annals of Statistics},
}

@book{hastie2009elements,
    author    = {Hastie and Tibshirani and Friedman and Friedman},
    year      = {2009},
    title     = {The Elements of Statistical Learning},
    publisher = {Springer},
}

@article{houlsby2011bayesian,
    author  = {Houlsby and Husz{\'a}r and Ghahramani and Lengyel},
    year    = {2011},
    title   = {{Bayesian} active learning for classification and preference learning},
    journal = {arXiv},
}

@article{kirsch2019batchbald,
    author  = {Kirsch and van Amersfoort and Gal},
    year    = {2019},
    title   = {{BatchBALD}: efficient and diverse batch acquisition for deep {Bayesian} active learning},
    journal = {Conference on Neural Information Processing Systems},
}

@article{kirsch2022unifying,
    author  = {Kirsch and Gal},
    year    = {2022},
    title   = {Unifying approaches in active learning and active sampling via {Fisher} information and information-theoretic quantities},
    journal = {Transactions on Machine Learning Research},
}

@mastersthesis{krizhevsky2009learning,
    author = {Krizhevsky},
    year   = {2009},
    title  = {Learning multiple layers of features from tiny images},
    school = {University of Toronto},
}

@article{lecun1998gradientbased,
    author  = {LeCun and Bottou and Bengio and Haffner},
    year    = {1998},
    title   = {Gradient-based learning applied to document recognition},
    journal = {Proceedings of the IEEE},
}

@article{lewis1994sequential,
    author  = {Lewis and Gale},
    year    = {1994},
    title   = {A sequential algorithm for training text classifiers},
    journal = {ACM-SIGIR Conference on Research and Development in Information Retrieval},
}

@article{lindley1956measure,
    author  = {Lindley},
    year    = {1956},
    title   = {On a measure of the information provided by an experiment},
    journal = {Annals of Mathematical Statistics},
}

@article{osband2022evaluating,
    author  = {Osband and Wen and Asghari and Dwaracherla and Lu and {Van Roy}},
    year    = {2022},
    title   = {Evaluating high-order predictive distributions in deep learning},
    journal = {Conference on Uncertainty in Artificial Intelligence},
}

@article{osband2022neural,
    author  = {Osband and Wen and Asghari and Dwaracherla and Lu and Ibrahimi and Lawson and Hao and O'Donoghue and {Van Roy}},
    year    = {2022},
    title   = {The neural testbed: evaluating joint predictions},
    journal = {Conference on Neural Information Processing Systems},
}

@article{osband2023epistemic,
    author  = {Osband and Wen and Asghari and Dwaracherla and Ibrahimi and Lu and {Van Roy}},
    year    = {2023},
    title   = {Epistemic neural networks},
    journal = {Conference on Neural Information Processing Systems},
}

@article{osband2023fine,
    author  = {Osband and Asghari and {Van Roy} and McAleese and Aslanides and Irving},
    year    = {2023},
    title   = {Fine-tuning language models via epistemic neural networks},
    journal = {International Conference on Machine Learning},
}

@article{rainforth2024modern,
    author  = {Rainforth and Foster and Ivanova and {Bickford Smith}},
    year    = {2024},
    title   = {Modern {Bayesian} experimental design},
    journal = {Statistical Science},
}

@article{shannon1948mathematical,
    author  = {Shannon},
    year    = {1948},
    title   = {A mathematical theory of communication},
    journal = {The Bell System Technical Journal},
}

@article{wang2021beyond,
    author  = {Wang and Sun and Grosse},
    year    = {2021},
    title   = {Beyond marginal uncertainty: how accurately can {Bayesian} regression models estimate posterior predictive correlations?},
    journal = {International Conference on Artificial Intelligence and Statistics},
}

@article{wen2022from,
    author  = {Wen and Osband and Qin and Lu and Ibrahimi and Dwaracherla and Asghari and {Van Roy}},
    year    = {2022},
    title   = {From predictions to decisions: the importance of joint predictive distributions},
    journal = {arXiv},
}

\clearpage
\appendix

\section{Results for data-store size $m=250$}

\begin{figure*}[h]
    \centering
    \begin{subfigure}[b]{\textwidth}
        \centering
        \includegraphics[width=\linewidth]{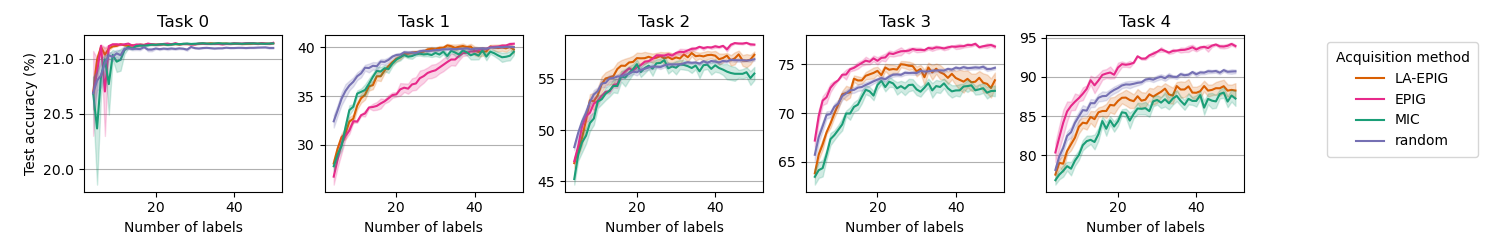}
        \caption{Dropout CNN on Split MNIST}
    \end{subfigure}
    \\
    \vspace{10pt}
    \begin{subfigure}{\textwidth}
        \centering
        \includegraphics[width=\linewidth]{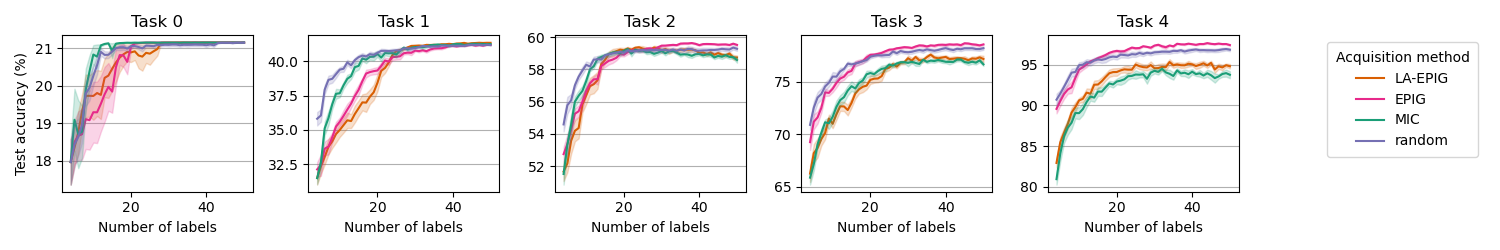}
        \caption{SimCLR encoder + dropout MLP on Split MNIST}
    \end{subfigure}
    \caption{
        Results for the setup presented in \Cref{fig:supervised_vs_semisupervised} except with data-store size $m=250$.
    }
    \label{fig:supervised_vs_semisupervised_m250}
\end{figure*}

\begin{figure*}[h]
    \centering
    \begin{subfigure}{\textwidth}
        \centering
        \includegraphics[width=\linewidth]{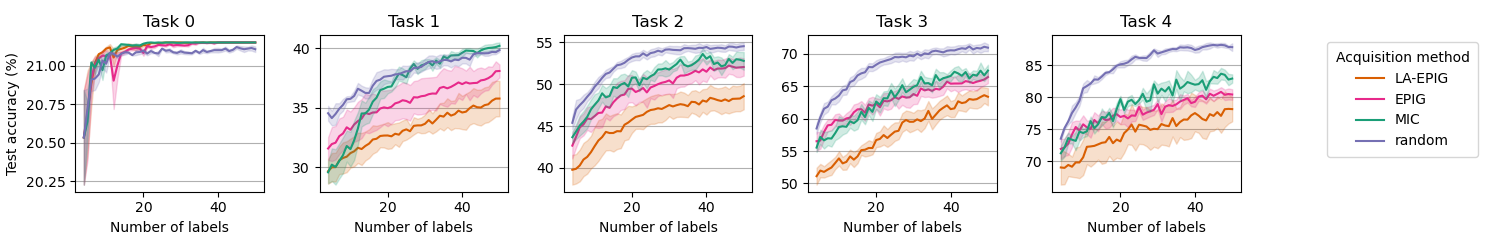}
        \caption{DINOv2 encoder + dropout MLP on Split MNIST}
    \end{subfigure}
    \\
    \vspace{10pt}
    \begin{subfigure}{\textwidth}
        \centering
        \includegraphics[width=\linewidth]{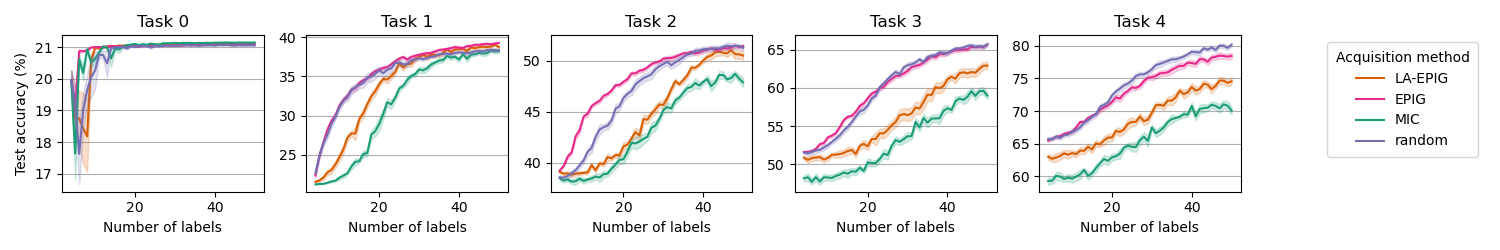}
        \caption{DINOv2 encoder + random forest on Split MNIST}
    \end{subfigure}
    \caption{
        Results for the setup presented in \Cref{fig:rf_vs_mlp_mnist} except with data-store size $m=250$.
    }
    \label{fig:rf_vs_mlp_mnist_m250}
\end{figure*}

\clearpage
\vspace*{0pt}

\begin{figure*}[h]
    \centering
    \begin{subfigure}{\textwidth}
        \centering
        \includegraphics[width=\linewidth]{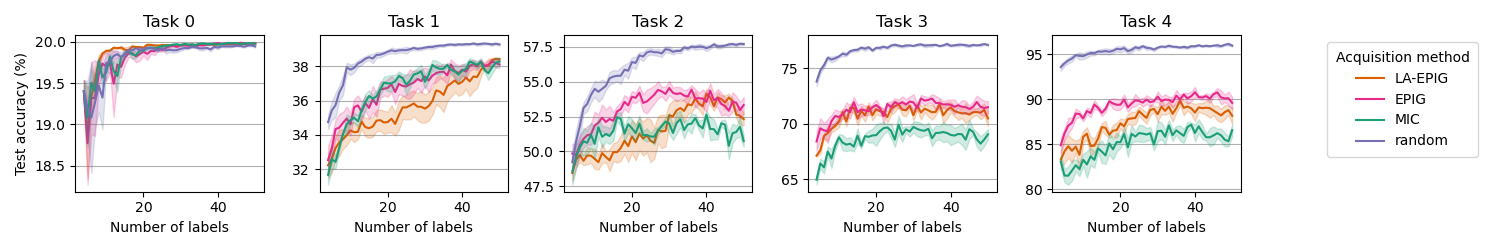}
        \caption{DINOv2 encoder + dropout MLP on Split CIFAR-10}
    \end{subfigure}
    \\
    \vspace{10pt}
    \begin{subfigure}{\textwidth}
        \centering
        \includegraphics[width=\linewidth]{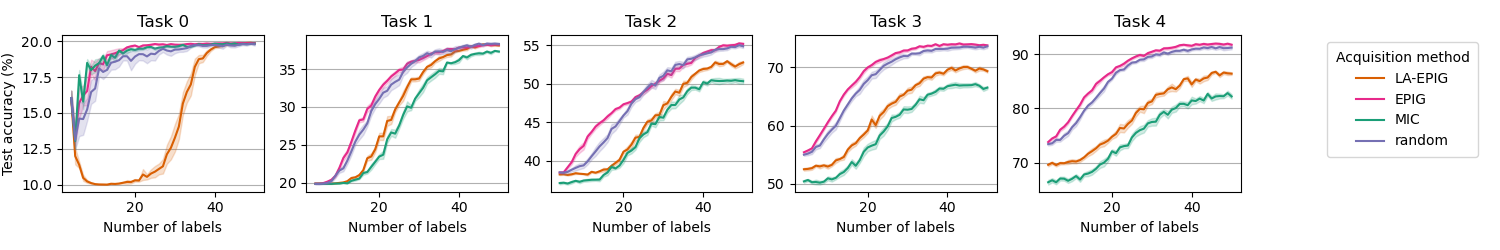}
        \caption{DINOv2 encoder + random forest on Split CIFAR-10}
    \end{subfigure}
    \caption{
        Results for the setup presented in \Cref{fig:rf_vs_mlp_cifar10} except with data-store size $m=250$.
    }
    \label{fig:rf_vs_mlp_cifar10_m250}
\end{figure*}

\clearpage

\section{Results for data-store size $m=500$}

\begin{figure*}[h]
    \centering
    \begin{subfigure}[b]{\textwidth}
        \centering
        \includegraphics[width=\linewidth]{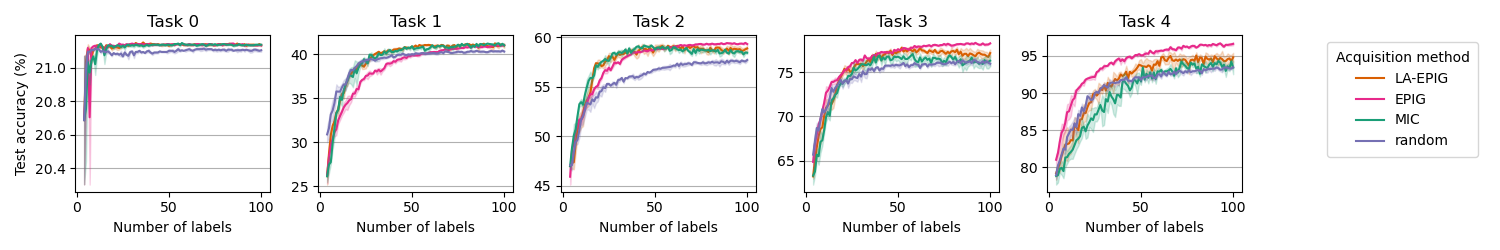}
        \caption{Dropout CNN on Split MNIST}
    \end{subfigure}
    \\
    \vspace{10pt}
    \begin{subfigure}{\textwidth}
        \centering
        \includegraphics[width=\linewidth]{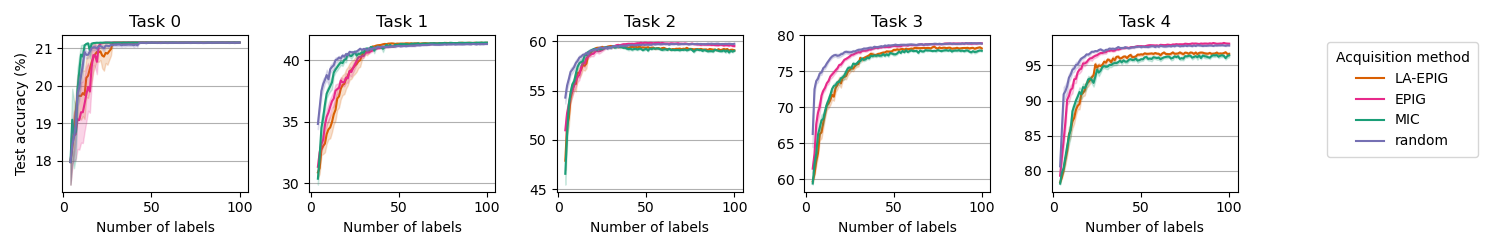}
        \caption{SimCLR encoder + dropout MLP on Split MNIST}
    \end{subfigure}
    \caption{
        Results for the setup presented in \Cref{fig:supervised_vs_semisupervised} except with data-store size $m=500$.
    }
    \label{fig:supervised_vs_semisupervised_m500}
\end{figure*}

\begin{figure*}[h]
    \centering
    \begin{subfigure}{\textwidth}
        \centering
        \includegraphics[width=\linewidth]{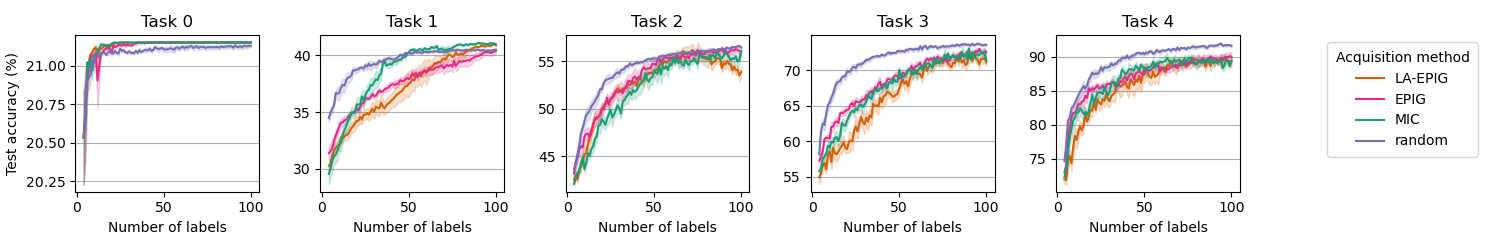}
        \caption{DINOv2 encoder + dropout MLP on Split MNIST}
    \end{subfigure}
    \\
    \vspace{10pt}
    \begin{subfigure}{\textwidth}
        \centering
        \includegraphics[width=\linewidth]{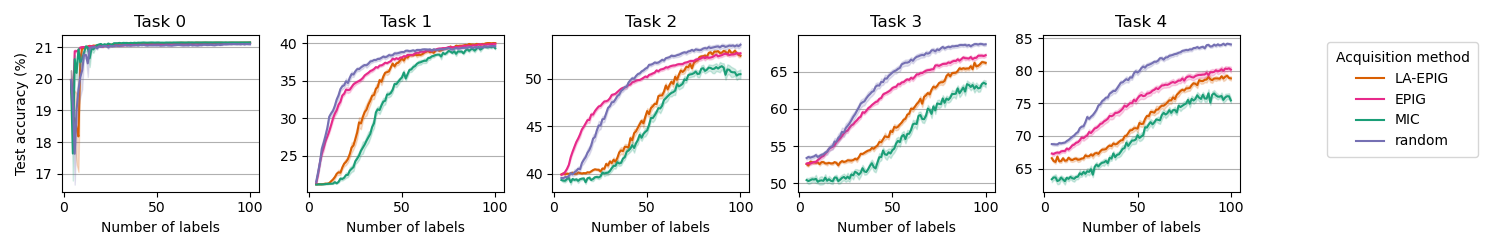}
        \caption{DINOv2 encoder + random forest on Split MNIST}
    \end{subfigure}
    \caption{
        Results for the setup presented in \Cref{fig:rf_vs_mlp_mnist} except with data-store size $m=500$.
    }
    \label{fig:rf_vs_mlp_mnist_m500}
\end{figure*}

\clearpage
\vspace*{0pt}

\begin{figure*}[t]
    \centering
    \begin{subfigure}{\textwidth}
        \centering
        \includegraphics[width=\linewidth]{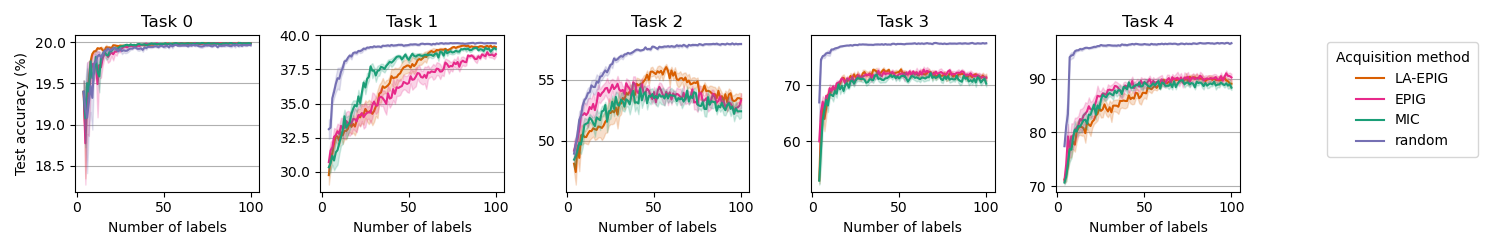}
        \caption{DINOv2 encoder + dropout MLP on Split CIFAR-10}
    \end{subfigure}
    \\
    \vspace{10pt}
    \begin{subfigure}{\textwidth}
        \centering
        \includegraphics[width=\linewidth]{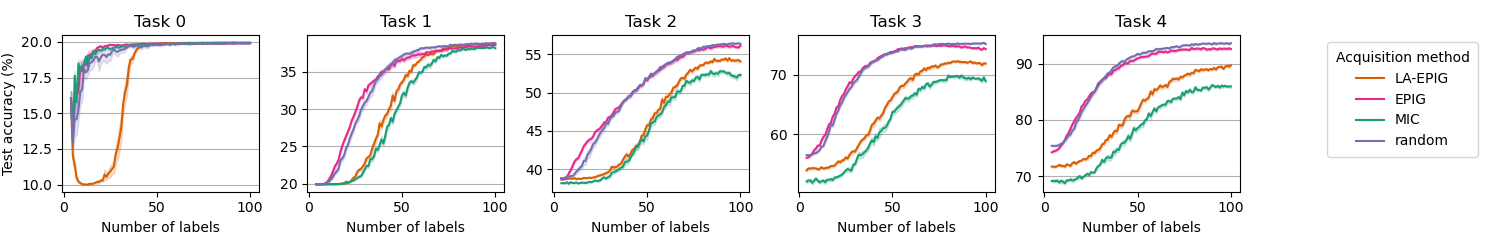}
        \caption{DINOv2 encoder + random forest on Split CIFAR-10}
    \end{subfigure}
    \caption{
        Results for the setup presented in \Cref{fig:rf_vs_mlp_cifar10} except with data-store size $m=500$.
    }
    \label{fig:rf_vs_mlp_cifar10_m500}
\end{figure*}

\end{document}